\documentclass[11pt]{article}

\usepackage[final]{acl}

\usepackage{times}
\usepackage{latexsym}
\usepackage[T1]{fontenc}
\usepackage[utf8]{inputenc}
\usepackage{microtype}
\IfFileExists{inconsolata.sty}{\usepackage{inconsolata}}{}
\usepackage{graphicx}
\usepackage{amsmath}
\usepackage{amssymb}
\usepackage{amsthm}
\usepackage{algorithm}
\usepackage{algorithmic}
\usepackage{booktabs}
\usepackage{multirow}
\usepackage{placeins}
\usepackage{colortbl}
\newcolumntype{L}[1]{>{\raggedright\arraybackslash}p{#1}}

\definecolor{headerbg}{RGB}{227,234,246}
\definecolor{groupbg}{RGB}{244,246,249}
\definecolor{oursbg}{RGB}{235,242,253}
\definecolor{bestbg}{RGB}{224,241,229}
\definecolor{alertbg}{RGB}{250,236,230}
\definecolor{rulegray}{RGB}{94,111,139}
\arrayrulecolor{rulegray}

\newcommand{\Ctr}{\textsc{CtR}}

\graphicspath{{figures/}}

\newcommand{\ind}{\mathbb{I}}
\newcommand{\E}{\mathbb{E}}
\newcommand{\system}{\textsc{UniJev}}
\newcommand{\Cora}{\textsc{CORA}}
\title{UniJev: Separating Accuracy from Calibration\\
in Trimodal System One Decision Models}

\author{
  \textbf{Jianing Wang}\textsuperscript{1},
  \textbf{Youwen Jian}\textsuperscript{1},
  \textbf{Derong Wang}\textsuperscript{1},
  \textbf{Jinyang Yu}\textsuperscript{1}, \\
  \textbf{Qingyi Si}\textsuperscript{\textbf{2}},
  \textbf{Chuanyu Qin}\textsuperscript{\textbf{3}},
  \textbf{Yanghua Xiao}\textsuperscript{\textbf{1,\textdagger}},
  \textbf{Hongcheng Guo}\textsuperscript{\textbf{1,\textdagger}} \\
  \textsuperscript{1}Fudan University \quad
  \textsuperscript{2}JD.com \quad
  \textsuperscript{3}Institute of Information Engineering, Chinese Academy of Sciences \\
  \textsuperscript{\textdagger}Corresponding authors \\
  \texttt{jnwang26@m.fudan.edu.cn}
}

\begin{document}
\maketitle

\begin{abstract}
System One models produce probabilistic judgments that software can consume directly, making reliable confidence as important as answer accuracy.
Yet joint training across text, images, and video remains underexplored, and the contributions of supervised fine-tuning (SFT) and reinforcement learning for calibrated decisions (RLCD) are difficult to separate.
We introduce \system{}, a trimodal framework that scores candidate answers with a shared head under joint supervised training across text, images, and video.
Supervised decision models improve accuracy over their backbone readouts across all three modalities, while the existing RLCD algorithms change accuracy little and have inconsistent effects on visual calibration across recipes.
These findings motivate a separate supervised objective for confidence calibration.
We introduce \Cora{} (Confidence--Outcome Reliability Alignment), which groups current predictions by confidence and optimizes each prediction's confidence toward the empirical accuracy of its bin.
At both 2B and 9B, \Cora{} lowers image and video expected calibration error (ECE) while preserving accuracy, without a reinforcement stage.
We release our code at \url{https://github.com/wjnwjnwj/UniJev}.
\end{abstract}

\begin{figure}[!t]
  \centering
  \includegraphics[width=.96\columnwidth]{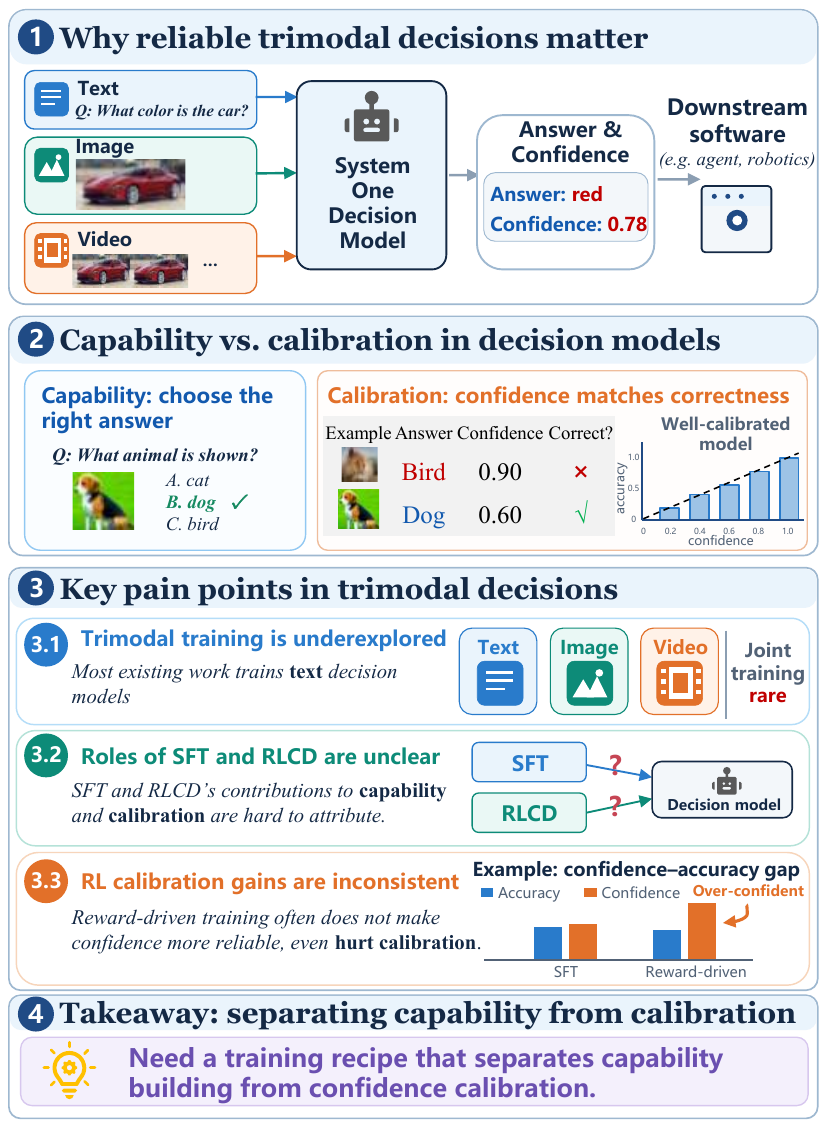}
  \caption{Motivation for reliable trimodal decisions. Joint trimodal training remains underexplored, and the effects of SFT and RLCD on accuracy and calibration are difficult to separate.}
  \label{fig:motivation}
\end{figure}
\section{Introduction}
\label{sec:intro}
\begin{table*}[!t]
\centering
\footnotesize
\setlength{\tabcolsep}{3pt}
\renewcommand{\arraystretch}{1.12}
\begin{tabular*}{\textwidth}{@{\extracolsep{\fill}}llll@{}}
\toprule
\rowcolor{headerbg}
\textbf{Method} & \textbf{Training signal} & \textbf{Estimator} & \textbf{Reference constraint} \\
\midrule
TypeSafe Jev & -- & -- & -- \\
Laya / Dohnuts & Log + spherical; RPS & Gaussian SF & Soft CE \\
Astro-RLCD & Log + spherical & Gaussian SF + LOO & None \\
Valen  & $c-(p_a-c)^2$ & Clipped SF + GN & KL + Brier \\
NanoJev & $2p_y-\lVert p\rVert^2$ & Paired SF + LOO & None \\
Decider RL & Return + belief log score & Clipped SF + direct & Retention KL \\
OpenJev-RLCD & Post-rationale Brier & Pathwise + rationale SF & KL \\
REINFORCE-A & CE + Brier & Analytic + outcome SF & KL \\
RLCR / Rewarding Doubt & Outcome + stated confidence & Sequence policy gradient & Recipe-specific \\
RLCD-lite & Correctness + Brier & Constrained GRPO & Recipe-specific \\
Von / OneJev / Open-Jev & CE + Brier & Direct gradient & None \\
Kev / NanoJev& CE & Direct gradient & None \\
GRPO / RLVR & Correctness $c$ & Categorical / sequence SF & KL \\
\midrule
\Cora{} (ours) & Current-group bin accuracy & Direct confidence gradient & None \\
\bottomrule
\end{tabular*}
\caption{Training designs of the methods compared in Table~\ref{tab:main}.
The reference constraint column lists each recipe's regularization toward a reference model or supervised objective, used to keep updates close to the initialization. 
$c$: sampled correctness; $p_a$: selected-answer probability; $y$: observed label; SF: score-function (REINFORCE-style) gradient; LOO: leave-one-out; GN: group normalization; RPS: ranked probability score. 
A listed recipe need not describe its released checkpoint; evaluation conventions and selected revisions appear in Appendix~\ref{sec:appendix-baselines}.}
\label{tab:related}
\end{table*}

System One models turn language-conditioned requests into judgments that software can consume directly.
Jev expresses these judgments through typed decisions and candidate probabilities \citep{typesafe2026jev,guo2026justaskjev}.
As the same workflow expands from documents to images and video, it requires both a capable decision model and reliable confidence.

Three challenges remain in developing a reliable training recipe.
First, joint training across text, images, and video remains underexplored, with many decision models focused on text \citep{laya2026,verdict2026,kev2026}.
Second, the respective contributions of supervised fine-tuning (SFT) and reinforcement learning for calibrated decisions (RLCD) \citep{typesafe2026jev,gao2026openjev} remain unclear.
Both stages update the same model, and some RLCD recipes combine reinforcement with additional supervision, making it difficult to attribute improvements to either component.
Third, rewarding correct decisions does not by itself ensure reliable confidence, because calibration requires the reported probabilities to agree with empirical correctness across questions.
Evaluating a training recipe therefore requires separate measurements of answer accuracy and confidence calibration.
Figure~\ref{fig:motivation} summarizes this distinction across the two training stages.

To address these challenges, we introduce \system{}, a trimodal decision model with separate training objectives for candidate selection and confidence calibration.
Text, images, and video share a multimodal backbone and a head that scores the candidate answers, producing answer distributions without autoregressive decoding.
Joint SFT trains candidate selection, followed by \Cora{} (Confidence--Outcome Reliability Alignment) for calibration.
\Cora{} groups current predictions by confidence and uses each sufficiently populated bin's empirical accuracy as a shared training target.
The groups and targets are recomputed as the model changes, and inference uses the trained model's probabilities directly.

We evaluate accuracy and calibration separately across text, image, and video.
Comparisons cover general-purpose multimodal models, released decision models (including Jev, OneJev, and Valen), and local training controls.
The local controls include SFT and reward-shaped RLCD (Shaped RLCD), as well as adaptations of GRPO \citep{shao2024deepseekmath}, OpenJev-RLCD/\Ctr{} \citep{gao2026openjev}, RLCD-lite \citep{rlcdlite2026}, and Rewarding Doubt \citep{stangel2025rewarding}; Appendix~\ref{sec:appendix-baselines} specifies these recipes.
On the common trimodal test suite, supervised decision models improve accuracy over backbone readouts, whereas the existing RLCD algorithms change accuracy little and have inconsistent effects on image and video calibration across recipes.
These results support treating visual calibration as a distinct training objective rather than assuming it improves alongside accuracy.
At both 2B and 9B, \Cora{} reduces image and video ECE while preserving accuracy, without a reinforcement stage.

Our contributions are threefold:
(1) a trimodal Jev-style model with a shared answer-scoring head and joint training across text, images, and video;
(2) controlled trimodal comparisons showing substantial accuracy gains after the supervised decision-model conversion but small accuracy changes and inconsistent visual-calibration effects after the existing RLCD algorithms;
and (3) \Cora{}, a supervised calibration method that updates the model using current confidence-group accuracy and improves image and video calibration in our experiments.

\section{Related Work}
\label{sec:related}

\paragraph{System One decision models.}
Jev provides typed probabilistic decisions without free-form generation \citep{typesafe2026jev,guo2026justaskjev}.
Text alternatives include encoder scoring (Laya, Verdict), option-conditioned attention (Jevlike), and pointer heads (Kev) \citep{laya2026,verdict2026,jevlike2026,kev2026}.
Decider uses option-token logits; NanoJev and Shen et al.'s OpenJev model candidate interactions \citep{decider2026,nanojev2026,shen2026openjev}.
Vision extensions include Laya Vision, Dohnuts, MoJev, Vision-Jev, and decider-vision; OneJev and Valen support text, images, and video \citep{layavision2026,dohnuts2026,mojev2026,visionjev2026,decider2026,onejev2026,valen2026}.
OmniJev combines decision heads with per-type calibration \citep{omnijev2026}.
\system{} studies joint trimodal training and confidence calibration through a shared answer-scoring head.

\paragraph{RLCD and confidence calibration.}
RLCD varies in reward, estimator, and supervision; Jev's full recipe remains unpublished \citep{typesafe2026jev,guo2026justaskjev}.
Examples include perturbed proper-score rewards (Laya, Dohnuts), updates without a reference-model constraint (Astro-RLCD), categorical rewards (Eve-RLCD), clipped updates (Valen), paired Brier estimates (NanoJev), and actor--belief learning (Decider) \citep{laya2026,dohnuts2026,astro2026rlcd,maio2026eve,valen2026,nanojev2026,decider2026}.
REINFORCE-A mixes analytic and sampled gradients; OpenJev-RLCD calibrates before rationale reinforcement \citep{shen2026openjev,gao2026openjev}.
RLCR and Rewarding Doubt reward stated confidence \citep{damani2025beyond,stangel2025rewarding}.
Proper scoring rules support probability learning, but scores also reflect predictive resolution beyond calibration \citep{brier1950verification,gneiting2007strictly,murphy1973new,vaicenavicius2019evaluating}.
Post-hoc methods include sigmoid scaling, histogram binning, and Bayesian binning \citep{platt2000probabilities,zadrozny2001obtaining,naeini2015obtaining}; calibration-error estimates depend on binning and sample size \citep{kumar2019verified,roelofs2022mitigating}.
\Cora{} uses current confidence-bin accuracy as detached supervision for the shared trimodal model, recomputing targets during training rather than applying a histogram map at inference.
Table~\ref{tab:related} summarizes recipes; evaluation and reproduction details appear in Appendix~\ref{sec:appendix-baselines}.

\section{Method}
\label{sec:method}

\begin{figure*}[t]
  \centering
  \includegraphics[width=\textwidth]{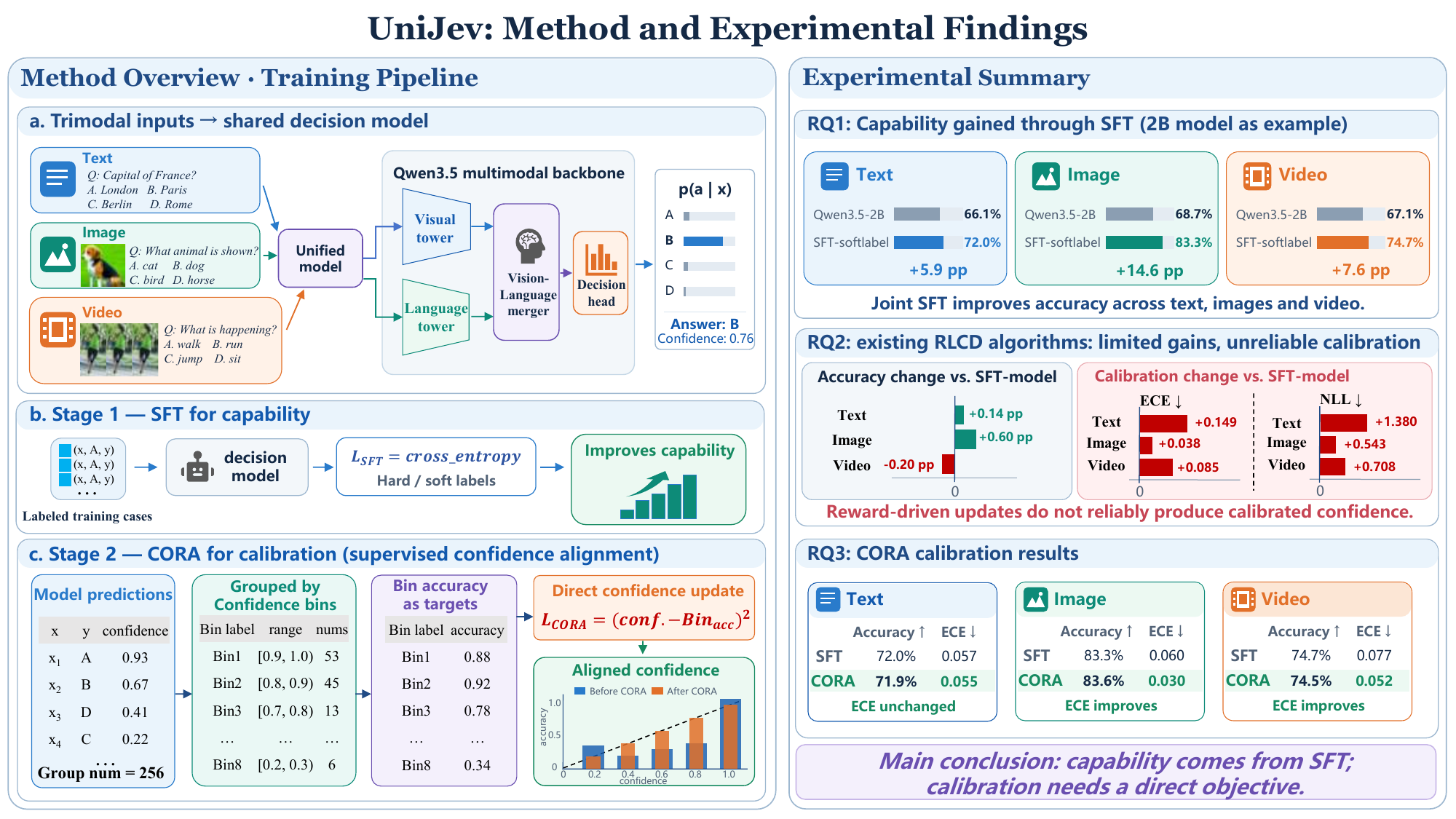}
  \caption{Overview of \system{} and the main findings.
  A shared trimodal decision model first learns candidate selection through the supervised conversion, then \Cora{} directly aligns confidence with the empirical accuracy of current confidence groups. 
  Most of the observed accuracy gain occurs during the supervised conversion; the evaluated reward-driven continuations do not consistently improve visual calibration; and \Cora{} lowers image and video ECE at both model scales while preserving accuracy.}
  \label{fig:overview}
\end{figure*}

\system{} uses a shared architecture for text, image, and video decisions.
SFT trains for first stage, and \Cora{} subsequently aligns confidence with empirical correctness across questions.
Figure~\ref{fig:overview} summarizes the shared architecture, the two training stages, and the main empirical findings.

\subsection{Shared trimodal architecture}
\label{sec:problem}
Each input $x$ contains a text or visual state, a question, and $K$ candidate answers with one correct option $y$.
We represent text and visual states in a common input format and score the supplied candidates with the same decision head.
The model returns $p_\theta(\cdot\mid x)\in\Delta^{K-1}$ in one forward pass and selects $\hat y_\theta(x)=\arg\max_a p_\theta(a\mid x)$.

A Qwen3.5 backbone encodes text and visual inputs \citep{qwen35models2026}.
Images and timestamped video frames pass through its visual tower.
We append the task, question, and candidate descriptions, followed by a fixed decision marker.
The hidden state at the end of each candidate represents that option, while the state at the decision marker summarizes the request and its candidates.
Two linear projections map these states to a shared dimension $d$, and each candidate's score is the scaled inner product of the two projected states:
\begin{equation}
\begin{aligned}
z_a &= \tfrac{1}{\sqrt d}
       \langle W_c h_{\mathrm{cand}(a)},W_d h_{\mathrm{dec}}\rangle,\\
p_\theta(a\mid x)&=\mathrm{softmax}(z)_a.
\end{aligned}
\label{eq:head}
\end{equation}
The head returns a candidate distribution directly, so inference does not require generating or parsing an answer string.

\subsection{Supervised candidate selection}
\label{sec:sft}
We train candidate selection jointly across the three modalities using labeled questions and their candidate answers.
For each question, the correct candidate receives target probability $\rho$, and the remaining mass is distributed uniformly among the alternatives.
All modalities use the same cross-entropy objective:
\begin{equation}
\begin{aligned}
q_y&=\rho,\qquad q_{a\ne y}=\frac{1-\rho}{K-1},\\
\mathcal L_{\mathrm{SFT}}&=-\sum_{a=1}^K q_a\log p_\theta(a\mid x).
\end{aligned}
\label{eq:sft}
\end{equation}
Setting $\rho=1$ gives hard-label supervision; $\rho<1$ applies label smoothing \citep{szegedy2016rethinking,mueller2019smoothing}.
The objective trains the shared head to assign higher probability to the correct option regardless of input modality.
This supervised stage provides the initialization for confidence calibration; Appendix~\ref{sec:appendix-hparams} lists the training settings.

\subsection{\Cora{}: supervised confidence calibration}
\label{sec:group-calibration}
\Cora{} (Confidence--Outcome Reliability Alignment) trains confidence against the observed correctness of predictions at similar confidence levels.
We estimate this relationship from groups of current predictions, with every group updating the same shared model.
For reported confidence $C$, the calibration condition is \citep{guo2017calibration,vaicenavicius2019evaluating}
\begin{equation}
\E\!\left[\ind\{\hat y_\theta(X)=Y\}\mid C=c\right]=c.
\label{eq:calibration}
\end{equation}
At update $t$, we draw a group $G_t$ of distinct questions from the trimodal training set $\mathcal D$.
The current model computes the selected answer, confidence, and correctness for each question:
\begin{equation}
\begin{aligned}
a_i^t&=\arg\max_a p_{\theta_t}(a\mid x_i),\\
 \bar c_i^t&=p_{\theta_t}(a_i^t\mid x_i),\quad
u_i^t=\ind\{a_i^t=y_i\}.
\end{aligned}
\label{eq:group-predictions}
\end{equation}
We partition $[0,1]$ into confidence intervals $I_b$ and estimate each bin's accuracy from the current group:
\begin{equation}
\begin{aligned}
B_{t,b}&=\{i\in G_t:\bar c_i^t\in I_b\},\\
k_{t,b}&=\frac{1}{|B_{t,b}|}\sum_{i\in B_{t,b}}u_i^t,
\qquad |B_{t,b}|>0.
\end{aligned}
\label{eq:group-target}
\end{equation}
For example, 15 correct predictions among 20 questions in $[0.8,0.9)$ give every question in that bin the target $0.75$, including its five errors.
The target $0.75$ estimates how often answers selected in this confidence range are correct.

Let $S_t$ contain questions whose bins have at least $m$ members.
We minimize
\begin{equation}
\begin{aligned}
\mathcal L_{\mathrm{CORA}}(\theta;\theta_t)
&=\frac{1}{|S_t|}\sum_{i\in S_t}\\
&\quad\left(p_\theta(a_i^t\mid x_i)-\operatorname{sg}[k_{t,b(i)}]\right)^2,
\end{aligned}
\label{eq:cora-loss}
\end{equation}
where the stop-gradient operator $\operatorname{sg}[\cdot]$ treats its argument as a constant during backpropagation.
The selected answers, bin assignments, and bin accuracies computed at $\theta_t$ are held fixed during the update; gradients at $\theta=\theta_t$ flow through the recomputed selected-answer probabilities.
The next group uses the updated model to recompute these quantities, as summarized in Algorithm~\ref{alg:cora-training}.
The support threshold $m$ excludes bins with too few observations from that update's loss.
If $S_t$ is empty, we skip the optimizer update and proceed to the next group.
\begin{algorithm}[t]
\caption{\Cora{}: supervised confidence calibration}
\label{alg:cora-training}
\small
\renewcommand{\algorithmicrequire}{\textbf{Input:}}
\begin{algorithmic}[1]
  \REQUIRE SFT parameters $\theta_0$; trimodal training set $\mathcal D$; intervals $\{I_b\}$; support $m$; group size $n$; updates $T$
  \FOR{$t=0,\dots,T-1$}
    \STATE sample $n$ distinct questions $G_t\subset\mathcal D$
    \STATE compute $a_i^t,\,\bar c_i^t,\,u_i^t$ for each $i\in G_t$ by \eqref{eq:group-predictions}
    \STATE form confidence bins $B_{t,b}$ and their accuracy targets $k_{t,b}$ by \eqref{eq:group-target}
    \STATE $S_t\leftarrow\{i:|B_{t,b(i)}|\geq m\}$
    \IF{$S_t\neq\varnothing$}
      \STATE recompute selected-answer probabilities and update the shared model using \eqref{eq:cora-loss}
    \ENDIF
  \ENDFOR
  \RETURN $\theta_T$
\end{algorithmic}
\end{algorithm}

Unlike post-hoc histogram binning \citep{zadrozny2001obtaining}, \Cora{} uses bins only to construct training targets.
At inference, the shared model returns a candidate-probability distribution, its argmax answer, and the maximum probability as confidence.
Because calibration updates can also change the selected answer, we evaluate accuracy alongside confidence calibration.
Section~\ref{sec:exp-setup} gives the training settings, and Appendix~\ref{sec:appendix-cora} describes the implementation.
\FloatBarrier

\section{Experiments}
\label{sec:experiments}
\begin{table*}[p]
\centering
\fontsize{8}{9.5}\selectfont
\setlength{\tabcolsep}{1.7pt}
\renewcommand{\arraystretch}{1.12}
\resizebox{\textwidth}{!}{%
\begin{tabular}{@{}l c|ccccc|ccccc|ccccc@{}}
\toprule
\rowcolor{headerbg}
& & \multicolumn{5}{c}{\textbf{Text} (764)} & \multicolumn{5}{c}{\textbf{Image} (1,858)} & \multicolumn{5}{c}{\textbf{Video} (502)} \\
\cmidrule(lr){3-7}\cmidrule(lr){8-12}\cmidrule(lr){13-17}
\rowcolor{headerbg}
\textbf{Model / method} & \textbf{B} & \textbf{Acc.} & \textbf{Conf.} & \textbf{ECE}$\downarrow$ & \textbf{NLL}$\downarrow$ & \textbf{Brier}$\downarrow$ & \textbf{Acc.} & \textbf{Conf.} & \textbf{ECE}$\downarrow$ & \textbf{NLL}$\downarrow$ & \textbf{Brier}$\downarrow$ & \textbf{Acc.} & \textbf{Conf.} & \textbf{ECE}$\downarrow$ & \textbf{NLL}$\downarrow$ & \textbf{Brier}$\downarrow$ \\
\midrule
\rowcolor{groupbg}
\multicolumn{17}{l}{\emph{General-purpose models}} \\
Qwen3.5-2B & 2 & 66.1 & 0.673 & 0.105 & 0.864 & 0.484 & 68.7 & 0.706 & 0.058 & 0.831 & 0.379 & 67.1 & 0.708 & 0.050 & 0.771 & 0.421 \\
Qwen3.5-9B & 9 & 76.8 & 0.828 & 0.060 & 0.587 & 0.322 & 76.7 & 0.811 & 0.043 & 0.637 & 0.296 & 73.3 & 0.748 & \textbf{0.031} & 0.586 & 0.327 \\
Qwen3.8-27B & 27 & 84.7 & 0.842 & \textbf{0.028} & \textbf{0.469} & 0.233 & 78.9 & 0.780 & 0.027 & 0.627 & 0.286 & 74.5 & 0.807 & 0.075 & 0.655 & 0.352 \\
InternVL3.5-8B & 8 & 76.4 & 0.893 & 0.132 & 0.823 & 0.365 & 83.0 & 0.881 & 0.053 & 0.516 & 0.229 & 71.5 & 0.798 & 0.084 & 0.703 & 0.369 \\
Gemma4-26B-A4B & 26 & 82.7 & 0.987 & 0.164 & 1.676 & 0.328 & 84.6 & 0.969 & 0.124 & 0.996 & 0.268 & 62.9 & 0.938 & 0.308 & 2.146 & 0.651 \\
\midrule
\rowcolor{groupbg}
\multicolumn{17}{l}{\emph{Released decision models}} \\
TypeSafe Jev & -- & \textbf{85.9} & 0.903 & 0.045 & 0.703 & \textbf{0.197} & -- & -- & -- & -- & -- & -- & -- & -- & -- & -- \\
Laya (English) & 0.421 & 63.4 & 0.745 & 0.112 & 0.935 & 0.476 & -- & -- & -- & -- & -- & -- & -- & -- & -- & -- \\
Laya (typed-dec.) & 0.421 & 62.0 & 0.624 & 0.063 & 0.851 & 0.465 & -- & -- & -- & -- & -- & -- & -- & -- & -- & -- \\
Laya Vision & 0.201 & 54.7 & 0.558 & 0.051 & 0.911 & 0.533 & 62.0 & 0.614 & 0.072 & 1.018 & 0.474 & -- & -- & -- & -- & -- \\
Von & 0.395 & 71.3 & 0.643 & 0.081 & 0.733 & 0.422 & -- & -- & -- & -- & -- & -- & -- & -- & -- & -- \\
Verdict & 0.151 & 47.1 & 0.576 & 0.116 & 0.968 & 0.573 & -- & -- & -- & -- & -- & -- & -- & -- & -- & -- \\
NIRNAY & 0.450 & 48.6 & 0.736 & 0.250 & 1.357 & 0.731 & -- & -- & -- & -- & -- & -- & -- & -- & -- & -- \\
Dohnuts & 0.8 & 67.1 & 0.743 & 0.107 & 0.809 & 0.458 & 76.0 & 0.703 & 0.062 & 0.690 & 0.316 & -- & -- & -- & -- & -- \\
OneJev & 9 & 79.3 & 0.793 & 0.034 & 0.518 & 0.279 & 86.3 & 0.853 & 0.021 & 0.379 & 0.183 & 73.5 & 0.735 & 0.061 & 0.653 & 0.358 \\
OmniJev & 4 & 69.6 & 0.845 & 0.149 & 0.878 & 0.429 & 81.6 & 0.857 & 0.042 & 0.498 & 0.232 & 63.1 & 0.761 & 0.129 & 0.938 & 0.480 \\
Open-Jev & 2 & 70.4 & 0.715 & 0.071 & 0.754 & 0.419 & -- & -- & -- & -- & -- & -- & -- & -- & -- & -- \\
Open-Jev & 9 & 79.7 & 0.810 & 0.071 & 0.634 & 0.317 & -- & -- & -- & -- & -- & -- & -- & -- & -- & -- \\
RLCR-hotpot & 7 & 61.8 & 0.668 & 0.070 & 0.835 & 0.496 & -- & -- & -- & -- & -- & -- & -- & -- & -- & -- \\
OpenJev & 4 & 79.5 & 0.871 & 0.077 & 0.612 & 0.287 & -- & -- & -- & -- & -- & -- & -- & -- & -- & -- \\
Astro-RLCD & 4 & 82.1 & 0.921 & 0.117 & 0.699 & 0.311 & -- & -- & -- & -- & -- & -- & -- & -- & -- & -- \\
Kev & 4 & 82.2 & 0.804 & 0.038 & 0.470 & 0.244 & -- & -- & -- & -- & -- & -- & -- & -- & -- & -- \\
decider & 2 & 75.1 & 0.808 & 0.059 & 0.550 & 0.330 & -- & -- & -- & -- & -- & -- & -- & -- & -- & -- \\
decider-vision & 2 & 74.9 & 0.839 & 0.095 & 0.669 & 0.368 & 80.2 & 0.831 & 0.030 & 0.533 & 0.245 & -- & -- & -- & -- & -- \\
NanoJev & 0.6 & 35.2 & 0.552 & 0.223 & 1.298 & 0.764 & -- & -- & -- & -- & -- & -- & -- & -- & -- & -- \\
Valen & 2 & 72.0 & 0.813 & 0.093 & 0.667 & 0.377 & \textbf{93.7} & 0.932 & \textbf{0.007} & \textbf{0.170} & \textbf{0.085} & 64.5 & 0.729 & 0.085 & 0.898 & 0.473 \\

\midrule
\rowcolor{groupbg}
\multicolumn{17}{l}{\emph{Supervised controls}} \\
SFT (2B) & 2 & 70.3 & 0.704 & 0.042 & 0.654 & 0.380 & 82.0 & 0.806 & 0.026 & 0.468 & 0.229 & 75.5 & 0.690 & 0.072 & 0.625 & 0.339 \\
SFT (9B) & 9 & 81.2 & 0.873 & 0.068 & 0.534 & 0.274 & 84.6 & 0.882 & 0.037 & 0.406 & 0.199 & \textbf{78.9} & 0.801 & 0.036 & \textbf{0.536} & \textbf{0.292} \\
SFT-softlabel (2B) & 2 & 72.0 & 0.737 & 0.057 & 0.693 & 0.387 & 83.3 & 0.779 & 0.060 & 0.482 & 0.218 & 74.7 & 0.681 & 0.077 & 0.648 & 0.344 \\
SFT-softlabel (9B) & 9 & 80.8 & 0.795 & 0.040 & 0.542 & 0.289 & 82.8 & 0.770 & 0.061 & 0.495 & 0.225 & 77.5 & 0.702 & 0.077 & 0.602 & 0.316 \\
\midrule
\rowcolor{groupbg}
\multicolumn{17}{l}{\emph{Local training baselines}} \\
Shaped RLCD & 2 & 72.4 & 0.940 & 0.220 & 1.641 & 0.475 & 84.2 & 0.941 & 0.099 & 0.748 & 0.250 & 74.1 & 0.914 & 0.173 & 1.048 & 0.409 \\
GRPO & 2 & 72.5 & 0.959 & 0.234 & 2.228 & 0.492 & 83.8 & 0.952 & 0.114 & 0.958 & 0.266 & 75.1 & 0.934 & 0.183 & 1.345 & 0.423 \\
CtR (OpenJev-RLCD) & 2 & 71.9 & 0.813 & 0.095 & 0.761 & 0.394 & 84.0 & 0.859 & 0.025 & 0.426 & 0.207 & 74.9 & 0.795 & 0.054 & 0.607 & 0.335 \\
RLCD-lite & 2 & 72.8 & 0.976 & 0.250 & 3.649 & 0.513 & 83.6 & 0.979 & 0.143 & 2.056 & 0.302 & 75.1 & 0.970 & 0.220 & 2.586 & 0.462 \\
Rewarding Doubt & 2 & 71.1 & 0.940 & 0.230 & 2.088 & 0.496 & 83.9 & 0.946 & 0.107 & 0.937 & 0.264 & 73.3 & 0.909 & 0.178 & 1.195 & 0.421 \\

\midrule
\rowcolor{groupbg}
\multicolumn{17}{l}{\system{} calibration training} \\
\rowcolor{oursbg}
\Cora{} (2B) & 2 & 71.9 & 0.767 & 0.055 & 0.643 & 0.400 & 83.6 & 0.851 & 0.030 & 0.449 & 0.216 & 74.5 & 0.767 & 0.052 & 0.620 & 0.342 \\
\rowcolor{oursbg}
\Cora{} (9B) & 9 & 81.0 & 0.836 & 0.048 & 0.541 & 0.288 & 82.6 & 0.807 & 0.031 & 0.468 & 0.220 & 77.5 & 0.736 & 0.063 & 0.579 & 0.310 \\
\bottomrule
\end{tabular}}
\caption{Model comparison on text (764 test questions), image (1,858), and video (502). 
Each modality reports accuracy (Acc., \%), mean maximum candidate probability (Conf.), 15-bin ECE, NLL, and multiclass Brier. 
The \Cora{} rows report the final checkpoints after 100 supervised calibration updates on the trimodal training set.
B denotes nominal parameter scale in billions.
-- denotes an unsupported modality or undisclosed scale. 
Released checkpoints use their native interfaces without local retraining. 
Our supervised controls use hard-label SFT at 2B/9B and soft-label SFT at 2B/9B. 
Local 2B recipe rows use a shared soft-label SFT initialization and a 500-step budget.}
\label{tab:main}
\end{table*}

We organize the experiments around three questions:
\begin{itemize}
    \item RQ1: Does joint SFT improve decision accuracy across text, images, and video?
    \item RQ2: What does RLCD add to an SFT model? better answers, more reliable confidence, or both?
    \item RQ3: How does confidence-bin supervision affect calibration, and which components account for the observed changes?
\end{itemize}

\subsection{Experimental setup}
\label{sec:exp-setup}
The training mixture combines text decisions from Kev's decision-v7 collection \citep{kev2026}, image decisions from Valen-Training-General-100k \citep{valen2026}, and video multiple-choice questions from the JeV video collection.
After source-group splitting, training contains 157{,}656 records, including 12{,}330 text, 98{,}142 image, and 47{,}184 video records. 
All training data cover 160{,}608 questions because some records contain multiple questions.
The main test suite has 3{,}124 questions, including 764 text, 1{,}858 image, and 502 video questions.
Training and test questions are disjoint by source group.

Table~\ref{tab:main} compares general-purpose models, released decision models, local RLCD adaptations, and hard- and soft-label SFT controls.
All training configurations, data-composition details, and reproduction protocols are provided in Appendices~\ref{sec:appendix-baselines}--\ref{sec:appendix-prompt}.

We use test-set answer accuracy as the measure of decision performance.
Confidence quality is measured by mean maximum candidate probability, 15-bin equal-width ECE \citep{naeini2015obtaining,guo2017calibration}, candidate-distribution NLL, and multiclass Brier score \citep{gneiting2007strictly,vaicenavicius2019evaluating}.
Appendix~\ref{sec:appendix-conf} examines its relationship to correctness through reliability diagrams and confidence-distribution statistics.

\subsection{Main results}
\label{sec:exp-main}
\paragraph{SFT improves decision accuracy across text, images, and video.}
Table~\ref{tab:main} shows that the supervised decision-model recipe improves accuracy across all three modalities.
At 2B, soft-label SFT reaches $72.0/83.3/74.7\%$ on text/image/video, compared with $66.1/68.7/67.1\%$ for Qwen3.5-2B.
At 9B, hard-label SFT reaches $81.2/84.6/78.9\%$, compared with $76.8/76.7/73.3\%$ for Qwen3.5-9B.
Soft-label SFT-9B reaches $80.8/82.8/77.5\%$, whose accuracy is lower than that of the hard-label model.
These comparisons assess the supervised conversion, including its decision head, rather than isolating the effect of the SFT objective alone.

\paragraph{The evaluated RLCD algorithms have heterogeneous calibration effects.}
Relative to their shared 2B SFT initialization, the five RLCD algorithms, including Shaped RLCD, GRPO, CtR, RLCD-lite, and Rewarding Doubt, change per-modality accuracy by $-1.4$ to $+0.9$ percentage points.
These differences are small compared with the gains from the supervised conversion.

Visual calibration differs more substantially across the RLCD algorithms.
CtR lowers image ECE from $0.060$ to $0.025$ and video ECE from $0.077$ to $0.054$.
Shaped RLCD, GRPO, RLCD-lite, and Rewarding Doubt increase ECE and NLL on both image and video: their confidence rises substantially while accuracy changes little, leaving them overconfident.

The existing RLCD algorithms therefore provide no consistent calibration benefit over SFT.
Appendix~\ref{sec:appendix-cases} illustrates the resulting predictions on individual text, image, and video questions.

\begin{table*}[t]
\centering
\footnotesize
\setlength{\tabcolsep}{4.5pt}
\begin{tabular}{@{}l rrr rrr rrr@{}}
\toprule
\rowcolor{headerbg}
& \multicolumn{3}{c}{\textbf{Text} (764)} & \multicolumn{3}{c}{\textbf{Image} (1,858)} & \multicolumn{3}{c}{\textbf{Video} (502)} \\
\cmidrule(lr){2-4}\cmidrule(lr){5-7}\cmidrule(lr){8-10}
\rowcolor{headerbg}
\textbf{Arm} & Acc. & Conf. & ECE$\downarrow$ & Acc. & Conf. & ECE$\downarrow$ & Acc. & Conf. & ECE$\downarrow$ \\
\midrule
\rowcolor{groupbg}
\multicolumn{10}{l}{\emph{Default}} \\
\Cora{} (10 bins, group 256, support 20) & 71.9 & 0.767 & \textbf{0.055} & 83.6 & 0.851 & 0.030 & 74.5 & 0.767 & 0.052 \\
\midrule
\rowcolor{groupbg}
\multicolumn{10}{l}{\emph{Primary single-factor contrasts}} \\
Bin accuracy $\to$ bin mean confidence & 71.9 & 0.989 & 0.274 & 83.0 & 0.992 & 0.162 & 74.7 & 0.986 & 0.240 \\
Confidence bins $\to$ one global target & 72.1 & 0.875 & 0.158 & 82.7 & 0.830 & 0.127 & 74.5 & 0.823 & 0.090 \\
Minimum support $20 \to 1$ & 72.3 & 0.802 & 0.095 & 83.6 & 0.844 & \textbf{0.026} & 75.3 & 0.763 & \textbf{0.045} \\
\midrule
\rowcolor{groupbg}
\multicolumn{10}{l}{\emph{Update scope}} \\
Trainable scope $\to$ head only & 72.0 & 0.802 & 0.083 & 83.5 & 0.848 & 0.032 & 74.9 & 0.761 & 0.061 \\
\midrule
\rowcolor{groupbg}
\multicolumn{10}{l}{\emph{Sensitivity}} \\
Bins $10 \to 5$ & \textbf{72.9} & 0.803 & 0.082 & 83.7 & 0.840 & 0.034 & 74.5 & 0.774 & 0.069 \\
Bins $10 \to 20$ & 71.6 & 0.828 & 0.112 & 83.2 & 0.866 & 0.036 & 74.3 & 0.792 & 0.074 \\
Group $256 \to 128$ & 71.9 & 0.868 & 0.149 & 83.0 & 0.881 & 0.054 & 75.3 & 0.824 & 0.095 \\
Group $256 \to 512$ & 71.6 & 0.796 & 0.080 & 83.6 & 0.840 & 0.031 & 74.3 & 0.752 & 0.058 \\
Minimum support $20 \to 10$ & 72.4 & 0.803 & 0.079 & \textbf{83.9} & 0.846 & 0.033 & \textbf{75.5} & 0.764 & 0.056 \\
Minimum support $20 \to 40$ & 71.9 & 0.872 & 0.154 & 83.0 & 0.879 & 0.055 & 74.5 & 0.834 & 0.094 \\
\bottomrule
\end{tabular}
\caption{\Cora{} ablations and sensitivity on the test suite.
Every arm shares the soft-label SFT initialization, training mixture, and 100-update protocol; each row changes one factor relative to the default and reports its final checkpoint. 
The default row is identical to the 2B \Cora{} row in Table~\ref{tab:main}.
}
\label{tab:cora-ablation}
\end{table*}

\subsection{\Cora{}: calibration results and ablations}
\label{sec:exp-position}

\paragraph{\Cora{} improves image and video calibration without sacrificing accuracy.}
In table~\ref{tab:main}, \Cora{}'s accuracy changes by $-0.1/{+}0.3/{-}0.2$ percentage points on text/image/video relative to soft-label SFT.
Image ECE falls from $0.060$ to $0.030$ and video ECE from $0.077$ to $0.052$, with lower NLL in both modalities.
Thus, direct confidence-bin supervision improves visual calibration while preserving decision accuracy.

\Cora{} has slightly higher image ECE than CtR ($0.030$ versus $0.025$) and lower video ECE ($0.052$ versus $0.054$).
Without a reinforcement stage, direct confidence-bin supervision yields lower image and video ECE and NLL than each of the four primarily reward-driven recipes.
Appendix~\ref{sec:appendix-conf} shows the remaining errors across confidence bins.

We also apply the recipe at 9B.
Starting from soft-label SFT-9B, \Cora{} (9B) reduces image ECE from $0.061$ to $0.031$ and video ECE from $0.077$ to $0.063$ while accuracy changes by at most $0.2$ points in every modality.
The consistent image and video reductions across both scales support a visual-calibration improvement.

Table~\ref{tab:cora-ablation} reports matched single-factor ablations.
Replacing bin accuracy with mean confidence removes correctness supervision, producing mean confidence near $0.99$ and ECE of $0.162$--$0.274$.
Replacing confidence-conditioned bins with one global correctness target also degrades all three modalities, including an increase in image ECE from $0.030$ to $0.127$.
Thus, both correctness supervision and confidence conditioning are important.
The remaining variants preserve accuracy within one percentage point of the default but expose differences between the two visual modalities.
Head-only updates nearly match image ECE ($0.032$ versus $0.030$) but trail on video, and the 512-question group is the closest sensitivity variant on both image and video.
Appendix~\ref{sec:appendix-cora} provides the procedural details.

\subsection{Confidence distributions behind Table~\ref{tab:main}}
\label{sec:exp-conf}
To examine the calibration differences in RQ2 and RQ3, we compare the test-confidence distributions behind Table~\ref{tab:main}.
Appendix~\ref{sec:appendix-conf} gives reliability diagrams for all rows (Figures~\ref{fig:rel-general}--\ref{fig:rel-local}) and a numeric profile (Table~\ref{tab:confidence-profile}).

Soft-label SFT places only $3.6\%$ of predictions above $0.95$ confidence.
Its overall accuracy is $0.792$ and mean confidence is $0.753$, but this aggregate underconfidence hides modality- and bin-specific behavior.
On image, about half of predictions fall in $[0.90,0.95)$, where empirical accuracy rounds to $1.00$; by contrast, the small text tail above $0.95$ is only $70\%$ correct.
The same model can therefore be underconfident on high-confidence image questions and overconfident on text questions.

Shaped RLCD, GRPO, RLCD-lite, and Rewarding Doubt place $69$--$93\%$ of predictions above $0.95$, with median confidence rounding to $1.000$ and a much larger Brier reliability term than SFT.
CtR and \Cora{} shift confidence less sharply.
Their image confidence--accuracy gaps are $+0.019$ and $+0.015$, compared with $-0.054$ for SFT.
This distributional view is consistent with \Cora{} correcting visual underconfidence.

At 9B, hard-label SFT is overconfident in every modality, whereas soft-label SFT is underconfident, consistent with label smoothing reducing confidence but not sufficient to isolate its effect.
Among released models, OneJev-9B has mean confidence--accuracy gaps within $\pm0.010$, showing that small aggregate gaps are attainable even though they do not guarantee per-bin calibration.

\FloatBarrier
\section{Conclusion}
\label{sec:conclusion}

\system{} unifies text, images, and video through a shared System One decision interface.
The supervised conversion improves accuracy across all three modalities, while the existing RLCD algorithms produce only small accuracy changes and inconsistent effects on visual calibration.
\Cora{} treats calibration as a separate supervised objective, aligning confidence with empirical correctness across questions while updating a shared model.
At both 2B and 9B, it improves image and video calibration while preserving accuracy, without a reinforcement stage.
These results support a division of labor between the two training stages.
The supervised conversion provides the observed accuracy gains, while direct confidence-bin supervision improves visual calibration.

\section*{Limitations}
\label{sec:limitations}

Our evidence for \Cora{}'s calibration benefit is limited to image and video decisions on the evaluated test suite and model scales.
Group-accuracy targets include each question's own outcome and may be noisy or transfer poorly beyond the training mixture.
Calibration updates can also change the model's answers.
Local RLCD comparisons use a 2B backbone and one seed per recipe, limiting conclusions about small accuracy differences.
Recipes that combine direct supervision with reinforcement do not isolate the contribution of the reinforcement stage.
Released-model pretraining and source overlap are uncontrolled, and ECE alone establishes neither subgroup calibration nor useful confidence ranking.
Our conclusions concern labeled, single-step decisions.

\bibliography{custom}

\clearpage
\appendix
\section{Baseline Evaluation and Reproduction}
\label{sec:appendix-baselines}

Released models retain native weights, preprocessing, and calibration, with adapters and decision heads loaded against their matching base models.
They are evaluated without local retraining.
Selected source revisions are Kev, Decider, Decider-Vision, NanoJev, and Valen.
NanoJev uses the step-400 CE release; Valen uses the SFT release with native 16-frame video processing and an FP32 decision head.
The baseline registry records checkpoint identifiers and source revisions.

Evaluation requires complete slices of 764 text, 1{,}858 image, and 502 video questions.
Each model receives all candidates in the prescribed order.
Serving-only candidate caps are removed while architectural limits remain; unsupported modalities are marked with dashes.
Verdict and OmniJev probabilities are conditioned on the supplied answers, with abstention mass archived separately.
Jev probability-mass errors up to 0.02 are renormalized; larger errors are rejected.
Confidence is the maximum candidate probability, and NLL uses a $10^{-12}$ floor.
Native media processing is retained, so released-model comparisons include differences in input processing and pretraining.

The local Shaped RLCD, GRPO, CtR, RLCD-lite, and Rewarding Doubt rows in Table~\ref{tab:main} share the same 2B soft-label SFT checkpoint.
Each continuation uses 500 steps on four GPUs with a global target of 32{,}768 input tokens per update.
Shaped RLCD uses reward shaping with confidence-shaped action rewards; GRPO uses correctness rewards with group normalization.
CtR adapts calibration-then-reinforcement to the categorical decision head with 250 steps of direct Brier calibration preceding 250 reinforcement steps, using the calibration checkpoint as reference.
It therefore tests the two-stage recipe on candidate selection rather than reproducing the source's rationale-generation experiments.
RLCD-lite jointly samples an answer from the decision distribution and a stated confidence from an auxiliary five-level confidence question, rewarding correctness minus the squared confidence--correctness gap with a group-mean baseline and no reference-model constraint.
Rewarding Doubt treats the proposed answer as fixed, asks an auxiliary eleven-level confidence question conditioned on that answer, and rewards the stated confidence's log score with an exact categorical expected-reward baseline and a Kullback--Leibler (KL) penalty (weight 0.05) toward the initialization.
Both adaptations update the shared model through the auxiliary confidence question; evaluation uses candidate probabilities, as for every row.

\Cora{} is evaluated as a model calibrated on the trimodal training set, using its own answers and candidate probabilities.
All local comparisons use the same test questions; every trained row reports its fixed-budget final checkpoint.

\section{Hyperparameters and Protocol Details}
\label{sec:appendix-hparams}

\paragraph{Data and evaluation protocol.}
The text tasks include classification, inference, and rating; the image tasks cover visual question answering, interfaces, documents, and games; and the video tasks cover event understanding and temporal reasoning.
Text evaluation uses Kev's transfer-v4 test split, with 444 Choice, 280 Noul, and 40 Score questions in 544 source groups.
Image and video test questions are randomly selected from their respective source pools, with source-group disjointness between training and test data.
Table~\ref{tab:main} evaluates the supervised controls at both 2B and 9B and five local 2B RLCD algorithms initialized from the same soft-label SFT model.
The shared prompt for general-purpose models appears in Appendix~\ref{sec:appendix-prompt}.

\paragraph{Configuration.}
We use the architecture in Section~\ref{sec:problem}, bfloat16 (BF16) backbone computation, 32-bit floating-point (FP32) scoring, and gradient checkpointing.
The decision projections have dimension $d=256$; low-rank adaptation (LoRA) \citep{hu2022lora} uses rank 32 and $\alpha=64$, trained together with the vision--language merger and decision head.
The decision marker is the literal suffix \texttt{Decision:}, with its final token supplying the decision readout state; training uses token-budget batching and reshuffles candidate order across epochs.
Inputs allow 32{,}768 tokens and up to 128 video frames, each capped at 65{,}536 pixels; over-length inputs are rejected.

\paragraph{Training.}
SFT-softlabel (2B) uses eight GPUs, 32{,}768 tokens per GPU per step, $\rho=0.9$, gradient clipping at 1.0, and seed 42.
Head/LoRA/merger learning rates are $(2\times10^{-4},5\times10^{-5},10^{-5})$.
Main-table 2B continuations use four GPUs, 500 steps, and a \emph{global} target of 32{,}768 tokens per update.
The soft-label SFT-9B control uses eight GPUs and 500 steps.

\paragraph{Calibration data.}
\Cora{} draws calibration groups from the trimodal SFT training set.
AdamW uses no weight decay, head/LoRA/merger rates $(2\times10^{-5},5\times10^{-6},10^{-6})$, clipping at 1, seed 42, and a fixed budget of 100 updates on four GPUs.
Each update groups 256 questions into 10 equal-width confidence bins with minimum support $m=20$; intervals are left-closed and right-open except for the final interval, which includes confidence $1.0$.
The reported model is the final checkpoint after 100 updates, without test-based checkpoint selection; test questions never contribute to training-bin statistics or gradients.

\begin{figure*}[t]
  \centering
  \includegraphics[width=\textwidth]{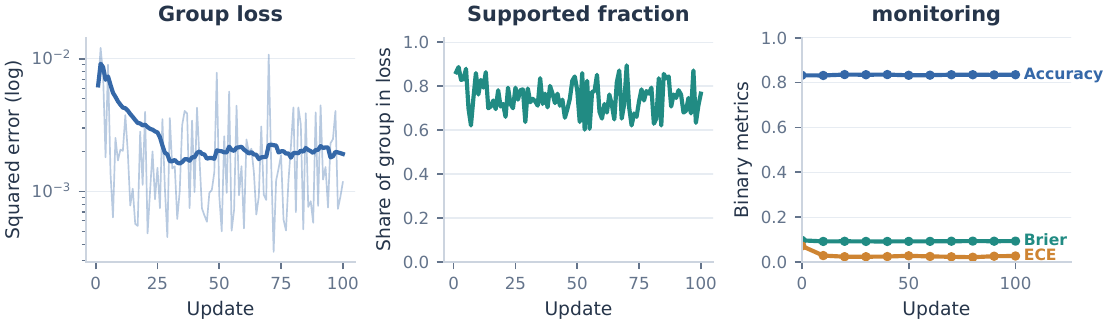}
  \caption{\Cora{} training dynamics over 100 updates. 
  Left: per-update group loss (thin) and its 25-update running mean (bold), shown on a logarithmic scale. 
  Center: fraction of each 256-question group included in the loss after applying the minimum-support rule. 
  Right: periodic monitoring accuracy, binary Brier score, and 15-bin ECE.}
  \label{fig:cora-training}
\end{figure*}

\section{CORA Procedure}
\label{sec:appendix-cora}

\Cora{} updates the SFT-trained model using statistics of its current predictions on training questions.
Calibration is trained into the model, and inference reads the trained probabilities directly.

\paragraph{Training and inference.}
The calibration groups come from the SFT trimodal training set.
For each optimizer update, compute the whole group's current predictions, pool confidence and correctness across workers, and form detached bin-accuracy targets.
To limit activation memory, collect statistics without gradients, then recompute each supported prediction at the same parameters and backpropagate its squared error.
Update parameters once per group; the next group uses fresh outputs.
If no bin reaches the minimum support, skip that optimizer update.
Algorithm~\ref{alg:cora-training} specifies this order.
At inference, return the trained model's argmax and candidate probabilities directly.

\paragraph{Training dynamics.}
Figure~\ref{fig:cora-training} summarizes the completed 100-update run.
The group loss decreases over training, while the supported fraction records how much of each group contributes to the calibration loss under the minimum-support rule.
Periodic accuracy, Brier, and ECE measurements provide additional optimization diagnostics.
These curves describe training dynamics only; the reported model remains the fixed-budget final checkpoint, and the test results in Tables~\ref{tab:main} and~\ref{tab:cora-ablation} are not used for checkpoint selection.

\paragraph{Ablation protocol.}
The single-factor arms in Table~\ref{tab:cora-ablation} share the soft-label SFT initialization, training mixture, question groups, trainable parameters, and 100-update budget unless the named factor changes them.
\label{sec:cora-ablation-protocol}
\label{sec:cora-sensitivity}
Appendix~\ref{sec:appendix-conf} examines the per-bin reliability of the resulting predictions.
Parameter updates can change predictions, so all studies track accuracy and answer changes rather than assuming preservation.

\section{Shared Prompt for General-Purpose Models}
\label{sec:appendix-prompt}
The original state and supported media precede the following common suffix. Candidate order is preserved for evaluation; labels expand with the actual candidate count, so no questions are dropped. Up to 26 options use A--Z; larger sets use fixed-width labels AA, AB, and so on (77 options end at CY).
\begin{quote}\footnotesize\ttfamily
Answer the question using the provided state and media.\\
Choose exactly one of the listed options.\\
Output only its label from: \{labels\}\\
Do not output an explanation or any other text.\\
\\
Question: \{question\}\\
Options:\\
\{options\}\\
\\
Answer:
\end{quote}
For a single-token label, its score is the corresponding vocabulary logit at the answer position; probabilities are normalized over the candidate labels. 
If a tokenizer splits a label, we use its full conditional sequence log-likelihood and normalize over labels, rather than scoring only its first token. 
Confidence is the maximum resulting candidate probability. 
Released decision models retain their native decision interfaces and released calibration.

\raggedbottom
\section{Cases from the Main Comparison}
\label{sec:appendix-cases}

We organize 22 cases by modality, which include eight text (T1--T8), eight image (I1--I8), and six video (V1--V6), one compact sheet per modality.
All cards compare Table~\ref{tab:main}'s soft-label 2B SFT initialization, the five 500-step RLCD continuations (Shaped RLCD, GRPO, RLCD-lite, Rewarding Doubt, and CtR), and the 100-update \Cora{} model.
Each card gives the input, candidates, annotation, and every model's predicted answer with its confidence percentage.
Display letters preserve candidate order; an asterisk marks an answer matching the annotation.
\Cora{} answers and confidence come from the same fixed-budget final checkpoint as Table~\ref{tab:main}, matched by question identifier.

Within each modality, cases cover shared correct answers, shared errors, and mixed outcomes.
Additional cases favor less represented task families and are then selected by a fixed record/question hash, with no repeated source group.
Selection is illustrative, not a representative estimate of calibration.
Confidence is rounded to one decimal percentage point; correctness follows dataset annotations.

\begin{figure*}[p]
  \centering
  \includegraphics[width=\textwidth]{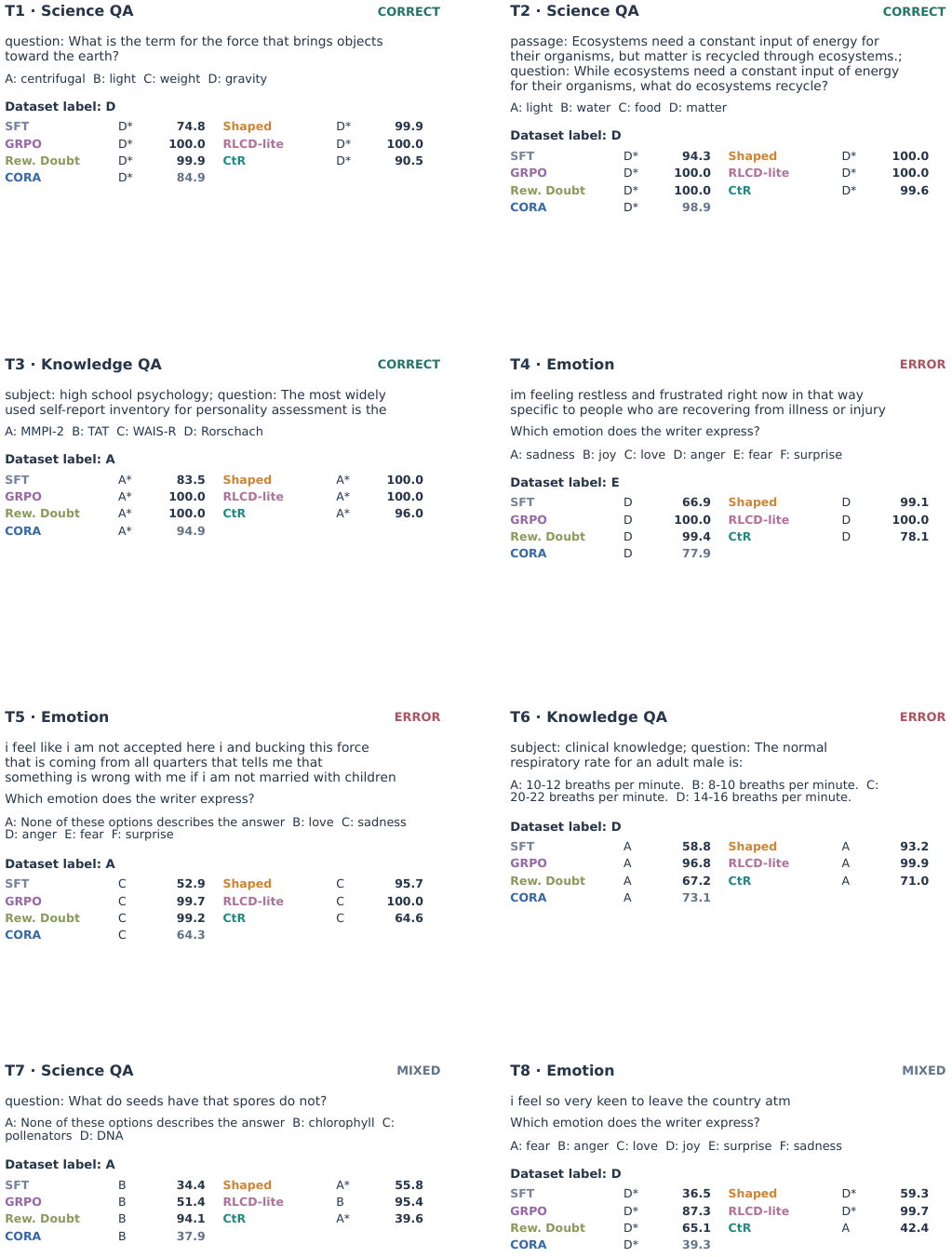}
  \caption{Text cases T1--T8: three shared correct answers, three shared errors, and two mixed outcomes across science, knowledge, and emotion questions. 
  Asterisks mark annotated answers; numbers are confidence percentages. 
  Candidate letters are display aliases, preserving the order and probabilities used in evaluation.}
  \label{fig:cases-text}
\end{figure*}

\paragraph{Text cases (Figure~\ref{fig:cases-text}).}
\label{sec:cases-text}
T1--T3 are shared correct answers across science and knowledge questions, T4--T6 shared errors, and T7--T8 mixed outcomes.
T1--T3 illustrate confidence increases on correct answers; T4--T6 show similar increases on errors.
T7 includes a none-of-the-listed-options answer selected correctly by Shaped RLCD and CtR, while T8 shows CtR changing SFT's correct emotion judgment.
These examples separate confidence changes from changes in answer correctness.

\begin{figure*}[p]
  \centering
  \includegraphics[width=.995\textwidth]{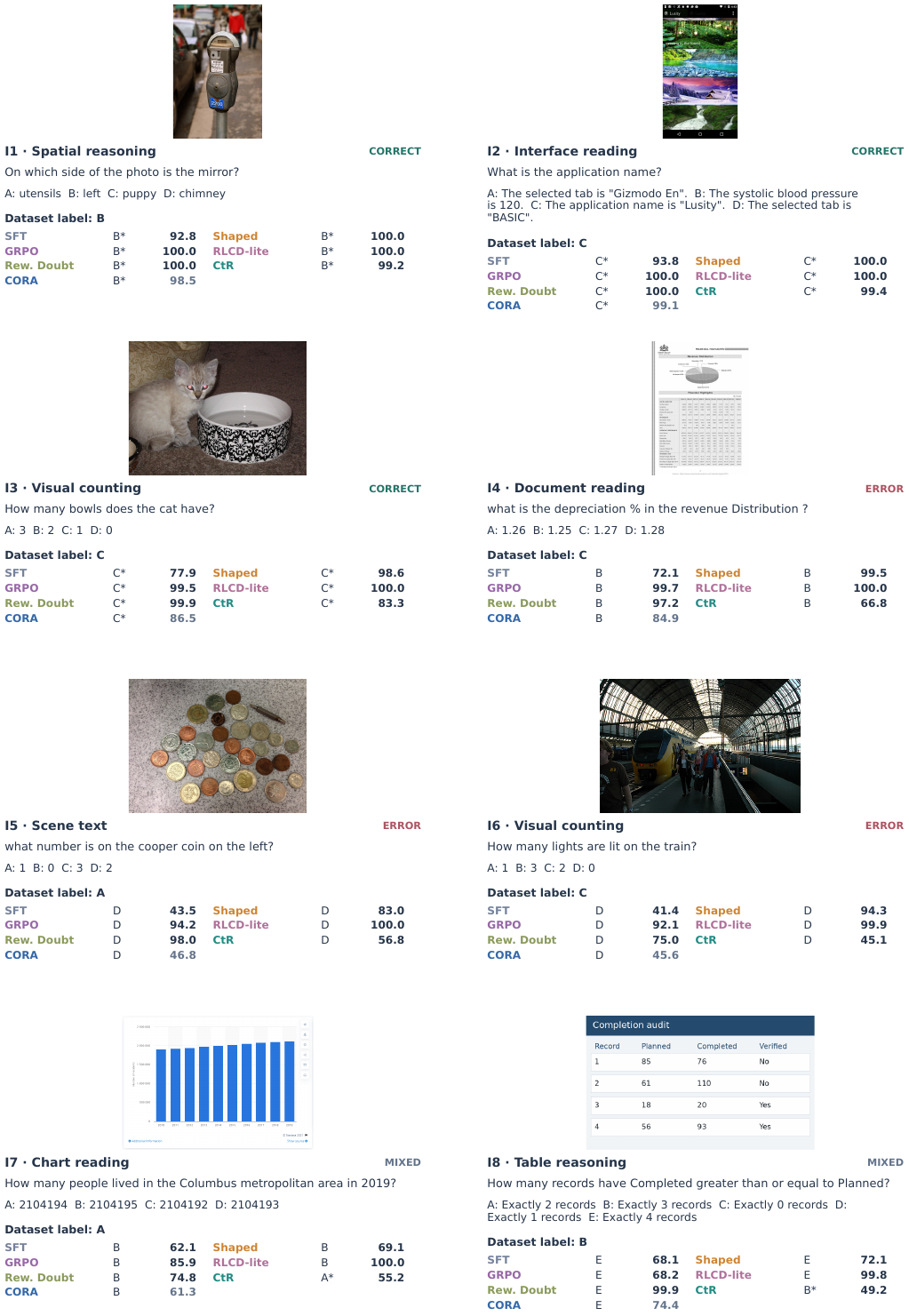}
  \caption{Image cases I1--I8: three shared correct answers, three shared errors, and two mixed outcomes across spatial, interface, counting, document, scene-text, chart, and table questions. 
  Predictions and confidences come from the same checkpoints as Table~\ref{tab:main}.}
  \label{fig:cases-image}
\end{figure*}

\paragraph{Image cases (Figure~\ref{fig:cases-image}).}
\label{sec:cases-image}
The image sheet covers spatial reasoning, interface reading, counting, document and scene text, charts, and tables.
I1--I3 are shared correct answers, I4--I6 shared errors, and I7--I8 mixed outcomes.
CtR selects the annotated answers in both mixed cases.
In I7, candidate populations differ by one person, beyond the chart's visible precision; correctness follows the annotation rather than demonstrating precise visual measurement.
All original images are shown without cropping.

\begin{figure*}[p]
  \centering
  \includegraphics[width=\textwidth]{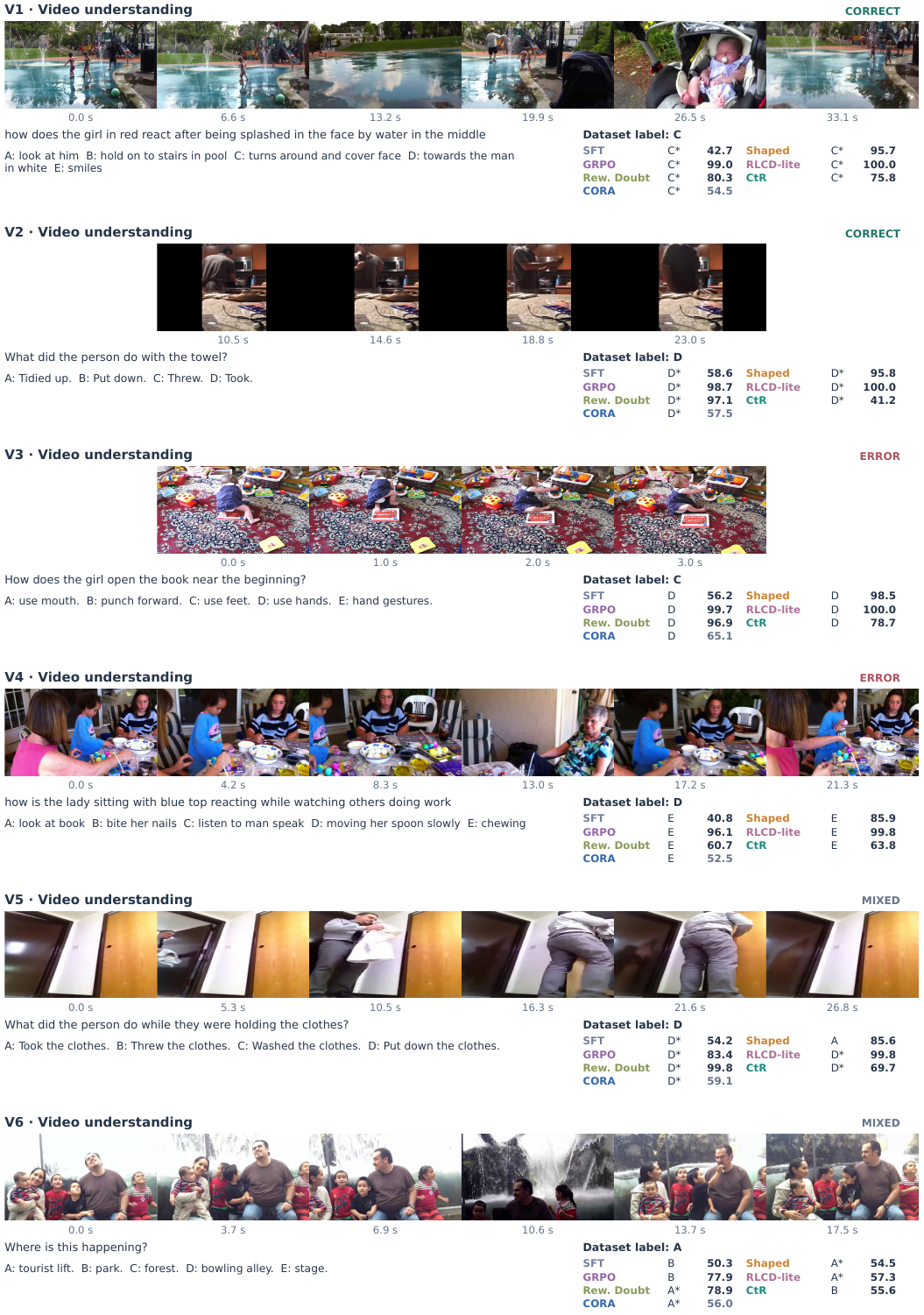}
  \caption{Video cases V1--V6: two shared correct answers, two shared errors, and two mixed outcomes for actions, reactions, and scene context. 
  Timestamps locate the displayed excerpts within each cached input.}
  \label{fig:cases-video}
\end{figure*}

\paragraph{Video cases (Figure~\ref{fig:cases-video}).}
\label{sec:cases-video}
V1--V2 are shared correct answers, V3--V4 shared errors, and V5--V6 mixed outcomes involving actions, reactions, and scene context.
CtR lowers confidence on the correct towel action (V2), yet raises it on the book-opening error (V3).
Shaped RLCD changes a correct answer to an error in V5 and an incorrect answer to the annotated answer in V6.
Each card shows a strip of up to six evenly spaced, uncropped cached frames with timestamps (fewer when the cached input contains fewer frames); these frames are excerpts, not the complete model input.
The cases illustrate changes in confidence and answer selection; aggregate calibration is reported in Tables~\ref{tab:main} and~\ref{tab:confidence-profile}.


\section{Reliability Profiles on the Test Suite}
\label{sec:appendix-conf}

\begin{figure*}[p]
  \centering
  \includegraphics[width=.92\textwidth]{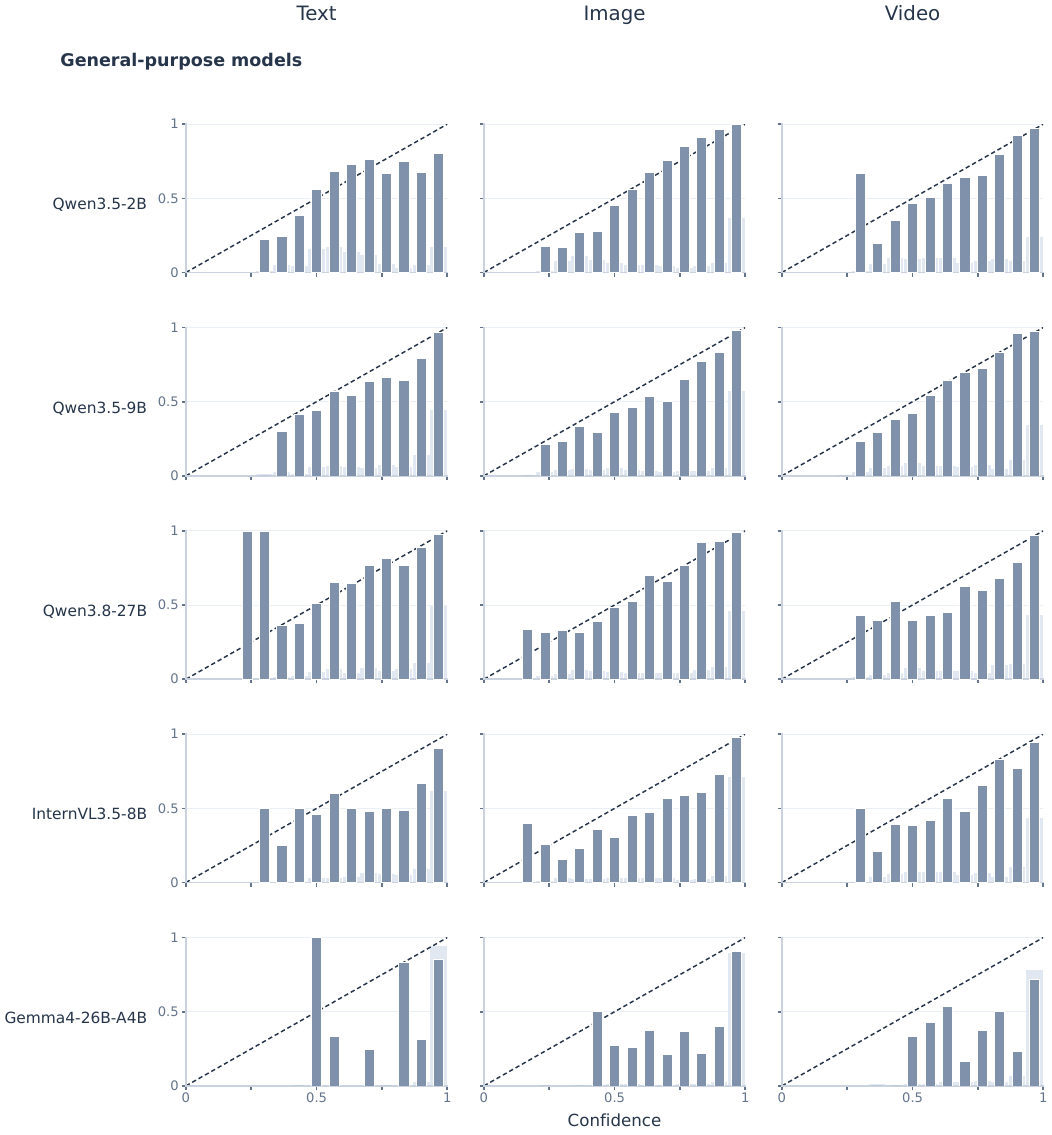}
  \caption{Per-bin reliability diagrams of the general-purpose models in Table~\ref{tab:main}, in table order. 
  Colored bars: empirical accuracy of each 15-bin confidence bin; gray bars: each bin's share of the slice's predictions; dashed diagonal: perfect calibration.}
  \label{fig:rel-general}
\end{figure*}

\begin{figure*}[p]
  \centering
  \includegraphics[width=.92\textwidth]{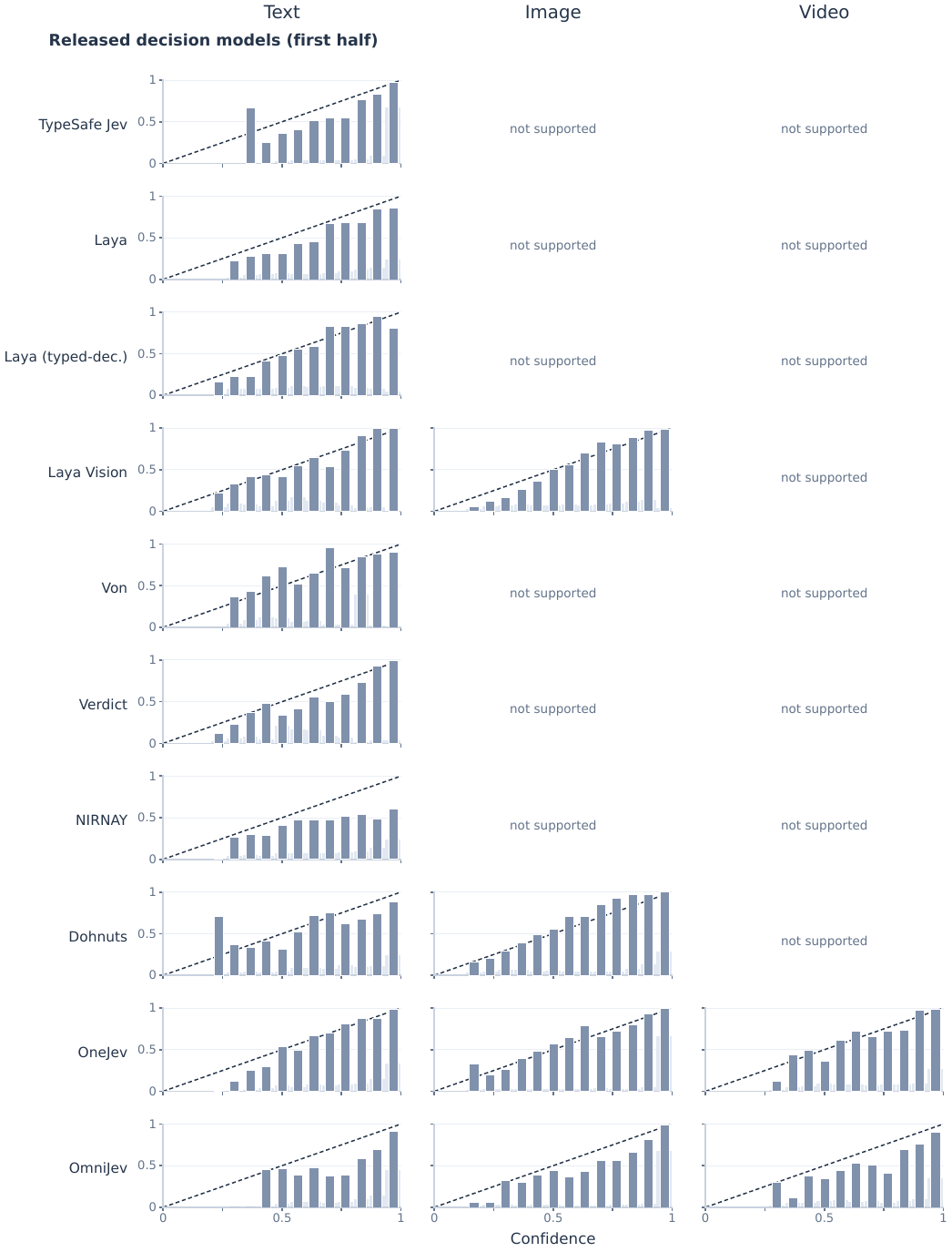}
  \caption{Per-bin reliability diagrams of the released decision models in Table~\ref{tab:main} (first half, table order). Axes and markings follow Figure~\ref{fig:rel-general}; modalities a model does not support are marked in place.}
  \label{fig:rel-released-a}
\end{figure*}

\begin{figure*}[p]
  \centering
  \includegraphics[width=.92\textwidth]{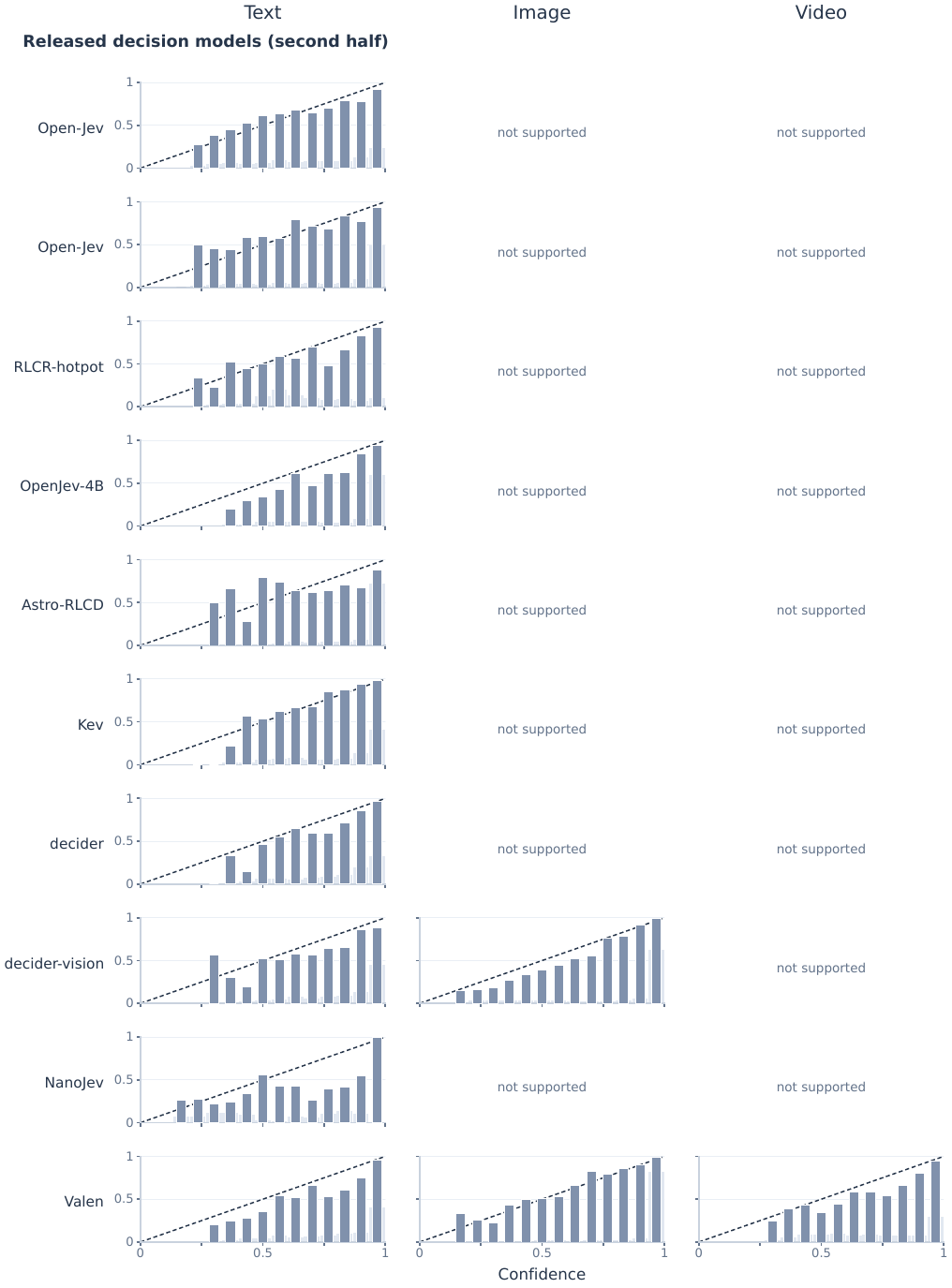}
  \caption{Per-bin reliability diagrams of the released decision models in Table~\ref{tab:main} (second half, table order). Axes and markings follow Figure~\ref{fig:rel-general}.}
  \label{fig:rel-released-b}
\end{figure*}

\begin{figure*}[p]
  \centering
  \includegraphics[width=.92\textwidth]{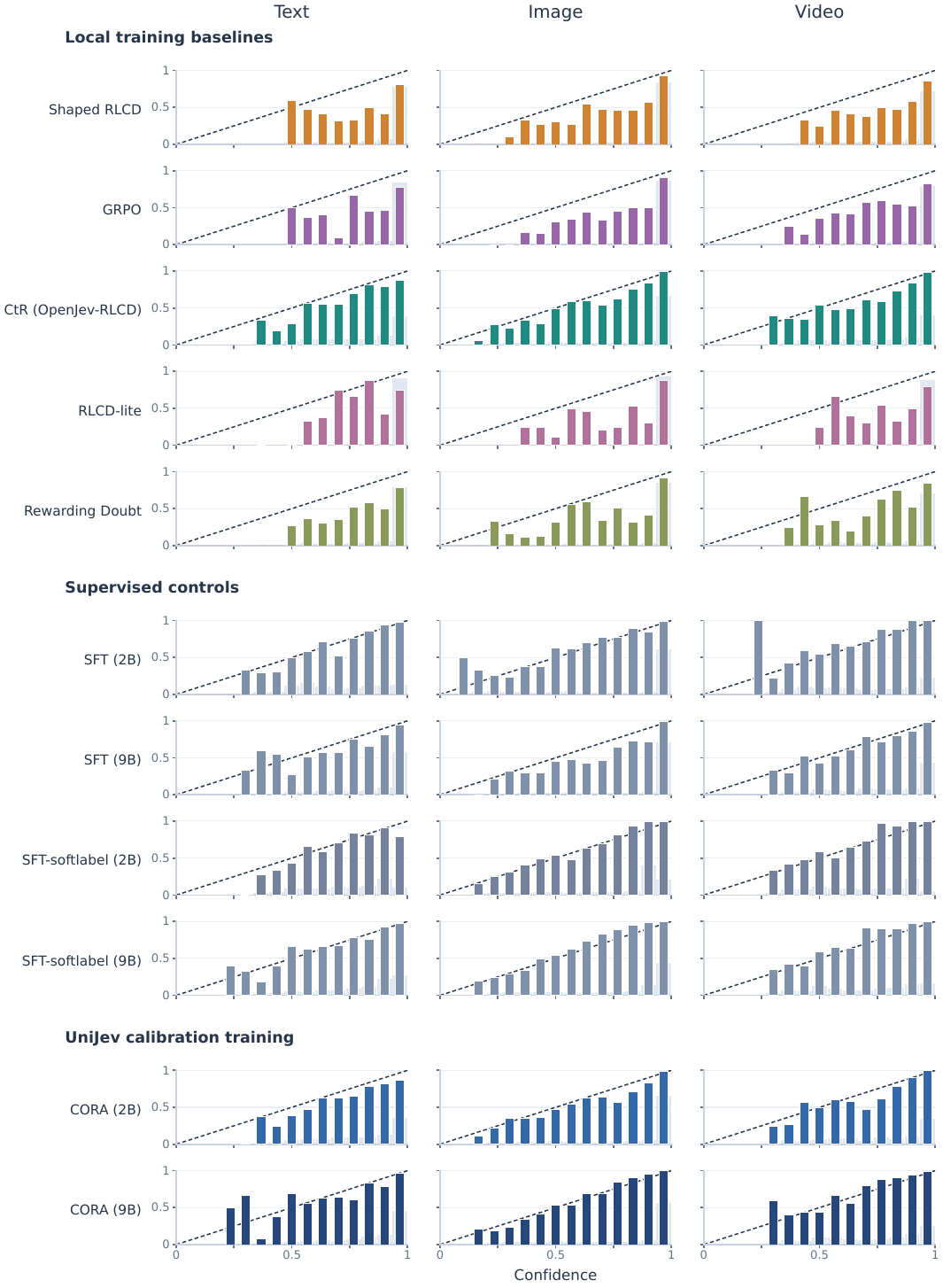}
  \caption{Per-bin reliability diagrams of the local training baselines, supervised controls, and \Cora{} (Table~\ref{tab:main} order). 
  The five RLCD baselines and \Cora{} (2B) share the soft-label SFT (2B) initialization, enabling a controlled comparison of the continuation recipes; \Cora{} (9B) starts from soft-label SFT (9B). 
  Axes and markings follow Figure~\ref{fig:rel-general}.}
  \label{fig:rel-local}
\end{figure*}

Figures~\ref{fig:rel-general}, \ref{fig:rel-released-a}, \ref{fig:rel-released-b}, and~\ref{fig:rel-local} give per-bin reliability diagrams for every row of Table~\ref{tab:main}, in the same order and grouping as the table.
In each panel the horizontal axis spans 15 equal-width confidence bins; the colored bars give the empirical accuracy of the predictions inside each bin, the light gray bars give each bin's share of the slice's predictions, and the dashed diagonal marks perfect calibration, where bin accuracy equals bin confidence.
Bars above the diagonal indicate underconfidence in that bin, bars below it overconfidence; empty positions are bins that received no predictions.
Modalities that a model does not support are marked in place, matching the dashes in Table~\ref{tab:main}.
Table~\ref{tab:confidence-profile} summarizes every distribution with four quantities per slice.
The signed gap is between mean confidence and accuracy, the share of predictions is above $0.95$, the reliability (calibration) term of the binary Brier decomposition \citep{murphy1973new} is with 15 equal-width bins, and the AUROC of confidence is against answer correctness.
AUROC measures whether confidence ranks correct predictions above errors; it does not measure agreement between confidence values and empirical accuracy.
Every displayed number is computed from the saved prediction files by the profile-rendering script, and each recomputed accuracy is checked against the corresponding Table~\ref{tab:main} cell.

The three regimes described in Section~\ref{sec:exp-conf} are directly visible among the locally trained rows of Figure~\ref{fig:rel-local}.
SFT-softlabel concentrates its mass below $0.95$ and sits above the diagonal at high confidence; on image, half of its predictions fall in $[0.90,0.95)$ where its empirical accuracy is $1.00$, so it is most underconfident exactly where it is most confident.
The four reward-driven recipes pile nearly all mass into the top bin while the accuracy of that bin stays far below the diagonal, the per-bin picture behind their inflated reliability terms.
CtR and \Cora{} retain graded confidence profiles and AUROC close to soft-label SFT. Shaped RLCD and GRPO change AUROC by at most $0.014$, whereas Rewarding Doubt and RLCD-lite reduce image AUROC from $0.934$ to $0.888$ and $0.847$, respectively.
Among the reference rows, the label-type contrast repeats at 9B (hard-label bins below the diagonal, soft-label bins above it in every modality), and the released models range from low top-label calibration error on this test suite (OneJev-9B) to nearly deterministic confidence (Gemma 4).

\begin{table*}[p]
\centering
\footnotesize
\setlength{\tabcolsep}{3.1pt}
\resizebox{\textwidth}{!}{%
\begin{tabular}{@{}lc|cccc|cccc|cccc@{}}
\toprule
\rowcolor{headerbg}
& & \multicolumn{4}{c}{\textbf{Text} (764)} & \multicolumn{4}{c}{\textbf{Image} (1,858)} & \multicolumn{4}{c}{\textbf{Video} (502)} \\
\cmidrule(lr){3-6}\cmidrule(lr){7-10}\cmidrule(lr){11-14}
\rowcolor{headerbg}
\textbf{Method} & \textbf{B} & $\Delta$ & ${>}0.95$ & REL & AUROC & $\Delta$ & ${>}0.95$ & REL & AUROC & $\Delta$ & ${>}0.95$ & REL & AUROC \\
\midrule
\rowcolor{groupbg}
\multicolumn{14}{l}{\emph{General-purpose models}} \\
Qwen3.5-2B & 2 & +0.012 & 13\% & 13.4 & 0.667 & +0.019 & 35\% & 5.5 & 0.913 & +0.037 & 20\% & 5.3 & 0.804 \\
Qwen3.5-9B & 9 & +0.060 & 41\% & 6.7 & 0.831 & +0.043 & 56\% & 4.2 & 0.916 & +0.015 & 31\% & \textbf{2.2} & \textbf{0.865} \\
Qwen3.8-27B & 27 & -0.004 & 46\% & 3.5 & 0.832 & -0.009 & 43\% & 1.4 & 0.900 & +0.062 & 39\% & 9.5 & 0.831 \\
InternVL3.5-8B & 8 & +0.129 & 58\% & 24.6 & 0.810 & +0.051 & 70\% & 7.1 & 0.942 & +0.083 & 40\% & 10.1 & 0.835 \\
Gemma4-26B-A4B & 26 & +0.160 & 94\% & 35.3 & 0.812 & +0.123 & 88\% & 26.2 & 0.930 & +0.308 & 75\% & 108.1 & 0.776 \\
\midrule
\rowcolor{groupbg}
\multicolumn{14}{l}{\emph{Released decision models}} \\
TypeSafe Jev & -- & +0.045 & 61\% & 5.8 & \textbf{0.883} & -- & -- & -- & -- & -- & -- & -- & -- \\
Laya (English) & 0.421 & +0.112 & 20\% & 14.6 & 0.761 & -- & -- & -- & -- & -- & -- & -- & -- \\
Laya (typed-dec.) & 0.421 & +0.003 & 3\% & 6.0 & 0.782 & -- & -- & -- & -- & -- & -- & -- & -- \\
Laya Vision & 0.201 & +0.011 & 0\% & 4.6 & 0.699 & -0.006 & 2\% & 6.8 & 0.865 & -- & -- & -- & -- \\
Von & 0.395 & -0.070 & 1\% & 12.8 & 0.699 & -- & -- & -- & -- & -- & -- & -- & -- \\
Verdict & 0.151 & +0.104 & 1\% & 17.0 & 0.677 & -- & -- & -- & -- & -- & -- & -- & -- \\
NIRNAY & 0.450 & +0.250 & 19\% & 78.3 & 0.611 & -- & -- & -- & -- & -- & -- & -- & -- \\
Dohnuts & 0.8 & +0.072 & 22\% & 15.1 & 0.711 & -0.058 & 24\% & 6.3 & 0.916 & -- & -- & -- & -- \\
OneJev & 9 & \textbf{+0.000} & 30\% & \textbf{2.3} & 0.849 & -0.010 & 65\% & 1.5 & 0.944 & \textbf{+0.000} & 23\% & 5.5 & 0.814 \\
OmniJev & 4 & +0.149 & 41\% & 32.7 & 0.792 & +0.041 & 66\% & 5.9 & 0.945 & +0.129 & 33\% & 22.0 & 0.804 \\
Open-Jev & 2 & +0.011 & 20\% & 5.7 & 0.718 & -- & -- & -- & -- & -- & -- & -- & -- \\
Open-Jev & 9 & +0.013 & 46\% & 7.6 & 0.770 & -- & -- & -- & -- & -- & -- & -- & -- \\
RLCR-hotpot & 7 & +0.050 & 9\% & 11.4 & 0.651 & -- & -- & -- & -- & -- & -- & -- & -- \\
OpenJev & 4 & +0.077 & 56\% & 9.7 & 0.856 & -- & -- & -- & -- & -- & -- & -- & -- \\
Astro-RLCD & 4 & +0.101 & 68\% & 15.4 & 0.777 & -- & -- & -- & -- & -- & -- & -- & -- \\
Kev & 4 & -0.018 & 37\% & 3.4 & 0.865 & -- & -- & -- & -- & -- & -- & -- & -- \\
decider & 2 & +0.057 & 26\% & 8.0 & 0.811 & -- & -- & -- & -- & -- & -- & -- & -- \\
decider-vision & 2 & +0.090 & 39\% & 11.3 & 0.751 & +0.029 & 60\% & 3.1 & 0.945 & -- & -- & -- & -- \\
NanoJev & 0.6 & +0.200 & 0\% & 71.1 & 0.615 & -- & -- & -- & -- & -- & -- & -- & -- \\
Valen & 2 & +0.093 & 35\% & 14.9 & 0.818 & \textbf{-0.005} & 80\% & \textbf{0.5} & \textbf{0.964} & +0.084 & 26\% & 11.6 & 0.777 \\

\midrule
\rowcolor{groupbg}
\multicolumn{14}{l}{\emph{Supervised controls}} \\
SFT (2B) & 2 & +0.001 & 11\% & 4.1 & 0.777 & -0.015 & 59\% & 2.4 & 0.932 & -0.065 & 20\% & 8.5 & 0.824 \\
SFT (9B) & 9 & +0.062 & 54\% & 7.4 & 0.831 & +0.037 & 70\% & 5.1 & 0.943 & +0.012 & 40\% & \textbf{2.2} & 0.843 \\
SFT-softlabel (2B) & 2 & +0.018 & 5\% & 5.8 & 0.728 & -0.054 & 3\% & 4.2 & 0.934 & -0.066 & 2\% & 8.4 & 0.832 \\
SFT-softlabel (9B) & 9 & -0.012 & 10\% & 3.2 & 0.791 & -0.057 & 27\% & 4.6 & 0.937 & -0.073 & 11\% & 8.1 & 0.841 \\

\midrule
\rowcolor{groupbg}
\multicolumn{14}{l}{\emph{Local training baselines}} \\
Shaped RLCD & 2 & +0.216 & 75\% & 57.3 & 0.722 & +0.099 & 82\% & 15.6 & 0.920 & +0.173 & 69\% & 35.4 & 0.841 \\
GRPO & 2 & +0.234 & 82\% & 60.6 & 0.714 & +0.114 & 85\% & 19.0 & 0.923 & +0.183 & 77\% & 35.7 & 0.827 \\
CtR (OpenJev-RLCD) & 2 & +0.095 & 33\% & 11.9 & 0.737 & +0.019 & 65\% & 2.5 & 0.936 & +0.046 & 37\% & 6.1 & 0.844 \\
RLCD-lite & 2 & +0.248 & 90\% & 65.6 & 0.698 & +0.143 & 93\% & 27.3 & 0.847 & +0.219 & 88\% & 52.7 & 0.793 \\
Rewarding Doubt & 2 & +0.230 & 76\% & 55.5 & 0.723 & +0.107 & 83\% & 21.8 & 0.888 & +0.176 & 68\% & 37.9 & 0.819 \\

\midrule
\rowcolor{groupbg}
\multicolumn{14}{l}{\system{} calibration training} \\
\Cora{} (2B) & 2 & +0.048 & 30\% & 9.1 & 0.727 & +0.015 & 63\% & 2.6 & 0.931 & +0.022 & 31\% & 6.0 & 0.832 \\
\Cora{} (9B) & 9 & +0.026 & 39\% & 7.7 & 0.802 & -0.019 & 53\% & 1.3 & 0.939 & -0.039 & 21\% & 6.2 & 0.836 \\
\bottomrule
\end{tabular}}
\caption{Confidence profiles of every Table~\ref{tab:main} row, computed from the published test predictions, in Table~\ref{tab:main} order. 
B: nominal parameter scale in billions; $\Delta$: mean confidence minus accuracy (signed); ${>}0.95$: share of predictions with confidence above $0.95$; REL: the reliability (calibration) term of the binary Brier decomposition with 15 equal-width bins, $\times 10^{3}$; AUROC: area under the receiver operating characteristic curve for confidence against answer correctness; --: unsupported modality or undisclosed scale.}
\label{tab:confidence-profile}
\end{table*}

\FloatBarrier

\end{document}